\documentclass{article}

\PassOptionsToPackage{square,sort&compress,comma,numbers}{natbib}
\usepackage[final]{nips_2017}
\usepackage[utf8]{inputenc} % allow utf-8 input
\usepackage[T1]{fontenc}    % use 8-bit T1 fonts
\usepackage{hyperref}       % hyperlinks
\usepackage{url}            % simple URL typesetting
\usepackage{booktabs}       % professional-quality tables
\usepackage{amsfonts}       % blackboard math symbols
\usepackage{nicefrac}       % compact symbols for 1/2, etc.
\usepackage{microtype}      % microtypography
\usepackage{xcolor}
\usepackage{mysymbols}
\usepackage{graphicx}
\usepackage{wrapfig}

\title{C-VQA: A Compositional Split of the \\Visual Question Answering (VQA) v1.0 Dataset}

\author{Aishwarya Agrawal$^\ast$, Aniruddha Kembhavi$^\dagger$, Dhruv Batra$^\ddagger$, Devi Parikh$^\ddagger$ \\
$^\ast$Virginia Tech, $^\dagger$Allen Institute for Artificial Intelligence, $^\ddagger$Georgia Institute of Technology\\
\texttt{aish@vt.edu, anik@allenai.org, \{dbatra, parikh\}@gatech.edu}}

\begin{document}

\maketitle

\begin{abstract}
Visual Question Answering (VQA) has received a lot of attention over the past couple of years. A number of deep learning models have been proposed for this task. However, it has been shown \cite{vqa-ba,YinYang,goyal2016making,johnson2016clevr} that these models are heavily driven by superficial correlations in the training data and lack \emph{compositionality} -- the ability to answer questions about \emph{unseen compositions} of \emph{seen concepts}. This compositionality is desirable and central to intelligence. In this paper, we propose a new setting for Visual Question Answering where the test question-answer pairs are compositionally novel compared to training question-answer pairs. To facilitate developing models under this setting, we present a new compositional split of the VQA v1.0 \cite{VQA} dataset, which we call Compositional VQA (C-VQA). We analyze the distribution of questions and answers in the C-VQA splits. Finally, we evaluate several existing VQA models under this new setting and show that the performances of these models degrade by a significant amount compared to the original VQA setting.
\end{abstract}

\section{Introduction}
\label{sec:intro}

% Visual Question Answering is an Artificial Intelligence (AI) task where given an image and a natural language question about the image (\eg, ``What kind of store is this?'', ``How many people are waiting in the queue?'', ``Is it safe to cross the street?''), the machine's task is to automatically produce an accurate natural language answer (``bakery'', ``5'', ``yes''). 
Automatically answering questions about visual content is considered to be one of the holy grails of artificial intelligence research. Visual Question Answering (VQA) poses a rich set of challenges spanning various domains such as computer vision, natural language processing, knowledge representation and reasoning. VQA is a stepping stone to visually grounded dialog and intelligent agents \cite{visdial,visdial_rl,de2016guesswhat,DBLP:journals/corr/MostafazadehBDG17}. In the past couple of years, VQA has received a lot of attention. Various VQA datasets have been proposed by different groups \cite{VQA,krishna2016visual,zhu2016visual7w,geman2015visual,malinowski2014multi,gao2015you,ren2015exploring,goyal2016making,YinYang} and a number of deep-learning models have been developed   \cite{VQA,ChenWCGXN15,DBLP:journals/corr/YangHGDS15,DBLP:journals/corr/XuS15a,DBLP:journals/corr/JiangWPL15,DBLP:journals/corr/AndreasRDK15,DBLP:journals/corr/WangWSHD15,kanan,hieco,DBLP:journals/corr/AndreasRDK16,DBLP:journals/corr/ShihSH15,DBLP:journals/corr/kim15,fukui,han,DBLP:journals/corr/IlievskiYF16,DBLP:journals/corr/WuWSHD15,DBLP:journals/corr/XiongMS16,zhou,saito}. 

However, it has been shown that despite recent progress, today's VQA models are heavily driven by superficial correlations in the training data and lack compositionality \cite{vqa-ba,YinYang,goyal2016making,johnson2016clevr} -- the ability to answer questions about \emph{unseen compositions} of \emph{seen concepts}. For instance, a model is said to be compositional if it can correctly answer [``What \textcolor{gray}{color} are the \textcolor{orange}{safety} \textcolor{blue}{cones}?'', ``\textcolor{green}{green}''] without seeing this question-answer (QA) pair during training, but perhaps having seen [``What \textcolor{gray}{color} are the \textcolor{orange}{safety} \textcolor{blue}{cones}?'', ``orange''] and [``What \textcolor{gray}{color} are the plates?'', ``\textcolor{green}{green}''] during training. 

% For instance, a model which can correctly answer the question ``What color are the safety cones?'' whose correct answer is ``green'' without having seen any such instance during training but having seen instances such as ``What color are the safety cones''? with correct answer ``orange'' and ``What color is the plate?'' with correct answer ``green'' would be considered to be compositional (\figref{fig:teaser}).

\begin{figure}[h]
  \centering
  \includegraphics[width=\textwidth]{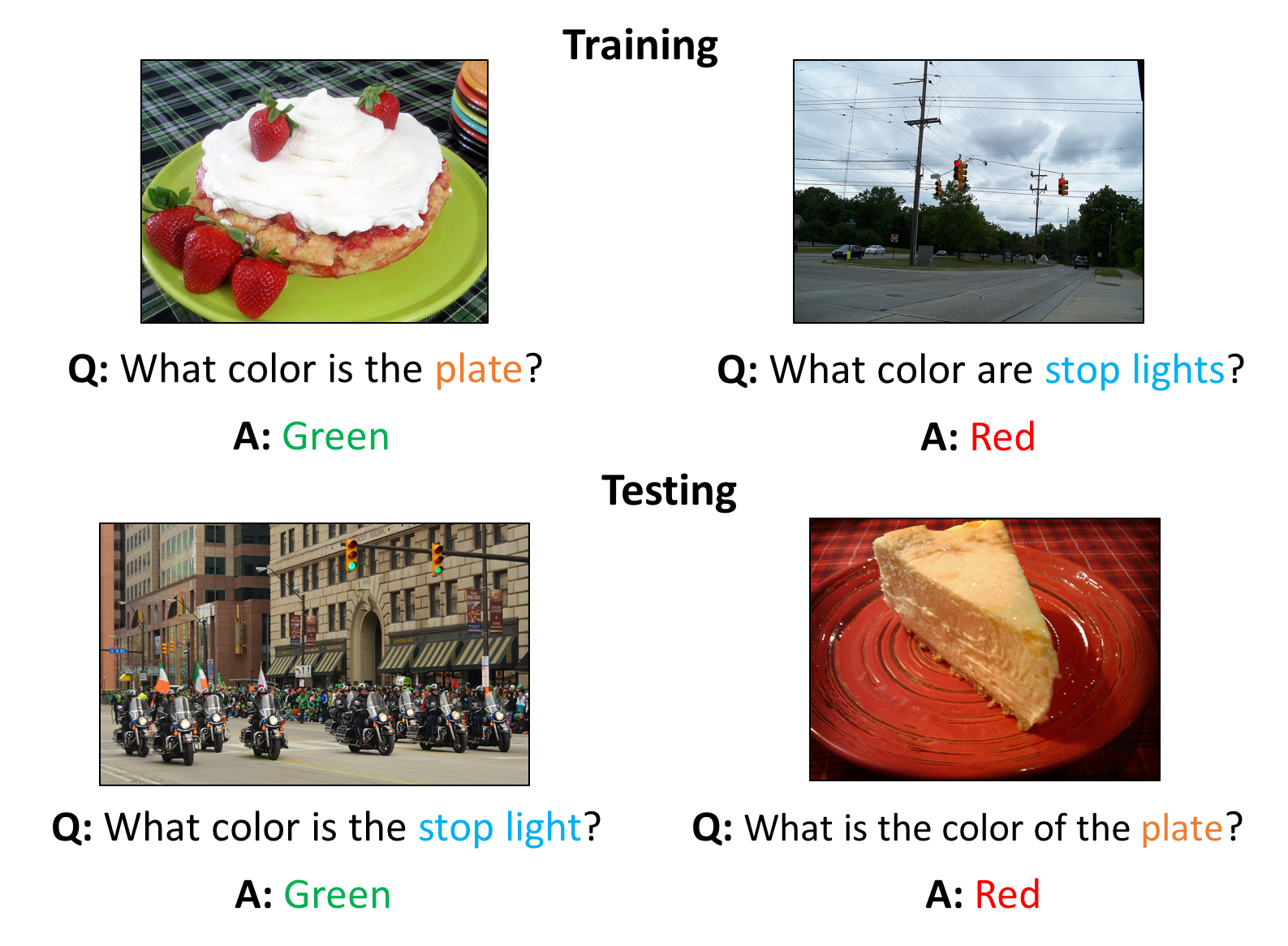}
  \caption{Examples from our Compositional VQA (C-VQA) splits. Words belonging to same concepts are highlighted with same color to show the training instances from which the model can learn those concepts.}
  \label{fig:teaser}
\end{figure}

% \begin{wrapfigure}[12]{R}{0.4\textwidth}
% \vspace{-15pt}
% \centering
% \includegraphics[width=0.4\textwidth]{figures/compositionality.png}
% \caption{Desired compositionality in VQA models. Words belonging to same concepts are highlighted with same color to show the training instances from which the model can learn those concepts.}
% \label{fig3}
% \end{wrapfigure}

In order to evaluate the extent to which existing VQA models are compositional, we create a compositional split of the VQA v1.0 dataset \cite{VQA}, called \textit{Compositional VQA (C-VQA)}. This new dataset is created by re-arranging the train and val splits of the VQA v1.0 dataset in such a way that the question-answer (QA) pairs in C-VQA test split are compositionally novel with respect to those in C-VQA train split, i.e., QA pairs in C-VQA test split are not present in C-VQA train splits but most concepts constituting the QA pairs in test split are present in the train split. \figref{fig:teaser} shows some examples from our C-VQA splits. Since, C-VQA test split contains the QA pair [``What is the color of the plate?'', ``red''], similar QA pairs such as [``What color is the plate?'', ``red''] are not present in C-VQA train split. But C-VQA train split contains other QA pairs consisting of the concepts ``plate'', ``red'' and ``color'' such as [``What color is the plate?'', ``green''] and [``What color are stop lights?'', ``red''].\footnote{It should be noted that in the VQA v1.0 splits, a given Image, Question, Answer (IQA) triplet is not shared across splits but a given QA pair could be shared across splits.} 

Evaluating a VQA model under such setting helps in testing -- 1) whether the model is capable of learning disentangled representations for different concepts (\eg, ``plate'', ``green'', ``stop light'', ``red''), 2) whether the model can compose the concepts learned during training to correctly answer questions about novel compositions at test time. Please see \secref{sec:dataset} for more details about C-VQA splits.

To demonstrate the difficulty of our C-VQA splits, we report the performance of several existing VQA models \cite{Lu2015, DBLP:journals/corr/AndreasRDK15,DBLP:journals/corr/YangHGDS15,hieco,fukui} on our C-VQA splits. Our experiments show that the performance of the VQA models drops significantly (with performance drop being smaller for models which are compositional by design such as the Neural Module Networks \cite{DBLP:journals/corr/AndreasRDK15}) when trained and evaluated on train and test splits (respectively) of C-VQA, compared to when these models are trained and evaluated on train and val splits (respectively) of the original VQA v1.0. Please see \secref{sec:baselines} for more details about these experiments.

\section{Related Work}
\label{sec:related_work}

\textbf{Visual Question Answering.} Several papers have proposed visual question answering datasets to train and test machines for the task of visual understanding \cite{VQA,krishna2016visual,zhu2016visual7w,gao2015you,malinowski2014multi,ren2015exploring,geman2015visual,YinYang,goyal2016making}. Over the span of time, the size of VQA datasets has become larger and questions have becomes more free form and open-ended. For instance, one of the earliest VQA datasets \cite{geman2015visual} considers questions generated using templates and consists of fixed vocabulary of objects, attributes, etc. \cite{malinowski2014multi} also consider questions whose answers come from a closed world. \cite{ren2015exploring} generate questions automatically using image captions and their answers belong to one of the following four types -- object, number, color, location. \cite{VQA,zhu2016visual7w,krishna2016visual,gao2015you} consist of free form open-ended questions. Of these datasets, the VQA v1.0 dataset \cite{VQA} has been used widely to train deep models. Performance of such models has increased steadily over the past two years on the test set of VQA v1.0 which has a similar distribution of data points as its training set. However, careful examination of the behaviors of such models reveals that these models are heavily driven by superficial correlations in the training data and lack compositionality \cite{vqa-ba}. This is partly because the training set of VQA v1.0 contains strong language priors which data-driven models can learn easily and can perform well on the test set which consists of similar priors as the training set, without truly understanding the visual content in images \cite{goyal2016making}, because it is easier to learn the biases of the data (or even our world) than to truly understand images. 

In order to counter the language priors, Goyal et al. \cite{goyal2016making} balance every question in the VQA v1.0 dataset by collecting complementary images for every question. Thus, for every question in the VQA v2.0 dataset, there are two similar images that have different answers to the question. Clearly, language priors are significantly weaker in the VQA v2.0 dataset. 
% They also show that the a number of state-of-art VQA models perform significantly worse on the VQA v2.0 dataset (compared to their performance on the VQA v1.0 dataset).
However, such balancing does not test for compositionality because the train and test distributions are similar. So, in order to test whether models can learn each concept individually irrespective of the correlations in the data and can perform well on a test set which has a different distribution of correlations compared to the training set, we propose a compositional split of the VQA v1.0 dataset, which we call Compositional-VQA (C-VQA).

\textbf{Compositionality.} The ability to generalize to novel compositions of concepts learned during training is desirable from any intelligent system. Compositionality has been studied in various forms in the vision community. Zero-shot object recognition using attributes is based on the idea of composing attributes to detect novel object categories \cite{lampert2009learning,jayaraman2014decorrelating}. More recently, \cite{atzmon2016learning} have studied compositionality in the domain of image captioning by focusing on structured representations (subject-relation-object triplets). We study compositionality for visual question answering where the questions and answers are open-ended and in free-form natural language. The work closest to us is \cite{johnson2016clevr} where they study compositionality in the domain of VQA. However, their dataset (images as well as questions) is synthetic and has only limited number of objects and attributes. On the contrary, our C-VQA splits consist of real images and questions (asked by humans) and hence involve a variety of objects and attributes, as well as activities, scenes, etc. Andreas et al. \cite{DBLP:journals/corr/AndreasRDK15,DBLP:journals/corr/AndreasRDK16} have developed compositional models for VQA that consist of different modules each specialized for a particular task. These modules can be composed together based on the question structure to create a model architecture for the given question. Although, compositional by design, these models have not been evaluated specifically for compositionality. Our C-VQA splits can be used to evaluate such models to test the degree of compositionality. In fact, we report the performance of Neural Module Networks on VQA v1.0 and C-VQA splits (\secref{sec:baselines}). 

The compositionality setting we are proposing is one type of zero-shot VQA where test QA pairs are novel. Other types of zero-shot VQA have also been explored. \cite{teney2016zero} propose a setting for VQA where the test questions (the question string itself or the multiple choices) contain atleast one unseen word. \cite{ramakrishnan2017empirical} propose answering questions about unknown objects (\eg, ``Is the dog black and white?'' where ``dog'' is never seen in training (neither in questions, nor in answers)).

\section{Compositional Visual Question Answering (C-VQA)}
\label{sec:dataset}

\subsection{C-VQA Creation}
\label{comp splits} 
The C-VQA splits are created by re-arranging the training and validation splits of the VQA v1.0 dataset \cite{VQA}\footnote{We can not use the test splits from VQA v1.0 because creation of C-VQA splits requires access to test annotations which are not publicly available.}. These splits are created such that the question-answer (QA) pairs in the C-VQA test split (\eg, Question: ``What color is the plate?'', Answer: ``green'') are not seen in the C-VQA train split, but in most cases, the concepts that compose the C-VQA test QA pairs (\eg, ``plate'', ``green'') have been seen in the C-VQA train split (\eg, Question: ``What color is the apple?'', Answer: ``Green'', Question: ``How many plates are on the table?'', Answer: ``4'').

 The C-VQA splits are created using the following procedure --

\textbf{Question Reduction:} Every question is reduced to a list of concepts needed to answer the question. For instance, \begin{center}``What color are the cones?'' is reduced to [``what'', ``color'', ``cone''].\end{center} We do this in order to reduce similar questions to the same form. For instance, ``What color are the cones?'' and ``What is the color of the cones?'' both get reduced to the same form -- [``what'', ``color'', ``cone'']'. This reduction is achieved using simple text processing such as removal of stop words and lemmatization.

% The details of the data processing involved in this step is provided in the supplement.

% \subsection{Data Pre-processing}
% \label{processing} 
% In order to convert the questions to their reduced forms, following pre-processing was performed on the questions -- 

% \begin{enumerate}
% \item \textbf{Part-of-speech (POS) tagging}: Words in the question were tagged with part-of-speech (e.g., noun, verb, adjective etc.) using Stanford POS tagger.

% \item \textbf{Filtering based on POS tagging}: Words corresponding to less important POS tags were filtered out (e.g., determiner, modal auxiliary, symbol etc.)

% \item \textbf{Stop-words removal}: Less important words were filtered out (e.g., all, being, yourselves, its etc.) with the help of NLTK stop-words list.

% \item \textbf{Lemmatization}: Every question word was reduced to its base form using wordnet lemmatizer from NLTK (e.g., cones -> cone)
% \end{enumerate}

\textbf{Reduced QA Grouping:} Questions having the same reduced form and the same ground truth answer are grouped together. For instance, \begin{center}[``What color are the cones?'', ``orange''] and [``What are the color of the cones?'', ``orange''] are grouped together whereas [``What color are the cones?'', ``green''] is put in a different group.\end{center} This grouping is done after merging the QA pairs from the VQA v1.0 train and val splits.

\textbf{Greedily Re-splitting:} A greedy approach is used to redistribute data points (image, question, answer) to the C-VQA train and test splits so as to maximize the coverage of the test concepts in the C-VQA train split while making sure QA pairs are not repeated between test and train splits. In this procedure, we loop through all the groups created above, and in every iteration, we add the current group to the C-VQA test split unless the group has already been assigned to the C-VQA train split. We always maintain a set of concepts\footnote{For a given group, concepts are the set of all unique words present in the reduced question and the ground truth answer belonging to that group} belonging to the groups in the C-VQA test split that have not yet been covered by the groups belonging to the C-VQA train split. From the groups that have not yet been assigned to either of the splits, we find the group that covers majority of the concepts (in the list) and add that group to the C-VQA train split.

The above approach results in 73.5\% of the unique C-VQA test concepts to be covered in the C-VQA train split. The coverage is 98.8\% when taking into account the frequency of occurrence of each concept in C-VQA test split.

\tableref{table:dataset} shows the number of questions, images and answers in the train and val splits of the VQA v1.0 dataset and those in train and test splits of the C-VQA dataset. We can see that the number of questions and the number of answers in the C-VQA splits is similar to that in the VQA v1.0 splits. However, the number of images in the C-VQA splits is more than that in the VQA v1.0 splits. This is because in the C-VQA splits, the same image can be present in both the train and the test sets. Note that there are three questions for every image in the VQA v1.0 dataset and the splitting for C-VQA is done based on QA pairs, not based on images. Consider the following two questions associated with the same image in VQA v1.0 -- ``What color are the cones?'' (with the answer ``orange'') and ``What time of day is it?'' (with the answer ``afternoon''). It is possible that ``What color are the cones?'' (along with the image and the ground-truth answers) gets assigned to C-VQA train split and ``What time of day is it?'' gets assigned to C-VQA test split. As a result, the image corresponding to these questions would be present in both the train and test splits of C-VQA.\footnote{To verify that sharing of images across splits does not make the problem easier, we randomly split the VQA v1.0 train+val into random-train and random-val. We then trained and evaluated the deeper LSTM Q + norm I model from \cite{VQA} on these new splits. We saw that the this new setup leads to only $\sim$1\% increase in the model performance compared to the VQA v1.0 train and val setup.}

\begin{table}[h]
\setlength{\tabcolsep}{3.2pt}
{\small
\begin{center}
\begin{tabular}{@{}llccc@{}}
\toprule
Dataset & Split & \#Questions & \#Images & \#Answers\\
%\hline
\midrule
    & Train & 248,349 & 82,783 & 2,483,490 \\
VQA (v1.0) \cite{VQA}   &  &  &  & \\
    & Val & 121,512 & 40,504 & 1,215,120 \\
%\hline
\midrule
    & Train & 246,574 & 118,663 & 2,465,740 \\
C-VQA (proposed)   &  &  &  & \\
    & Test & 123,287 & 86,700 & 1,232,870\\
\bottomrule
\end{tabular}
\end{center}
}
\caption {Statistics of the VQA v1.0 and our C-VQA splits.}
\label{table:dataset}
\end{table}

\subsection{C-VQA Analysis}
\label{analysis}

In this section we analyze how the distributions of questions and answers in the C-VQA train and test splits differ from those in the VQA v1.0 train and val splits.

\begin{figure*}[h]
\centering
\includegraphics[width=1\linewidth]{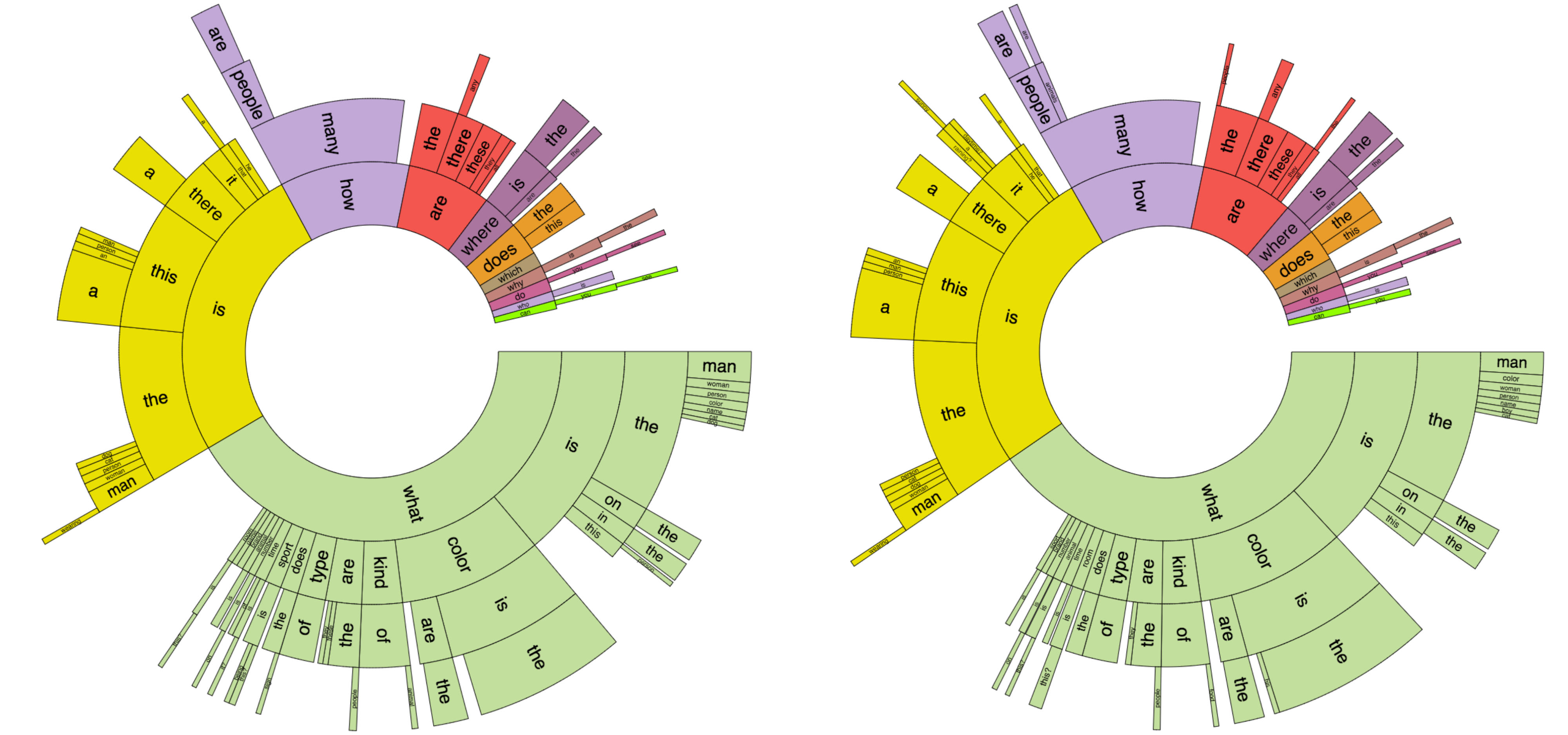}
\caption{Distribution of questions by their first four words for a random sample of 60K questions for \textbf{C-VQA train split} (left) and \textbf{C-VQA test split} (right). The ordering of the words starts towards the center and radiates outwards. The arc length is proportional to the number of questions containing the word. White areas are words with contributions too small to show.}
%\vspace{-5pt}
\label{fig:comp_ques}
\end{figure*}

\textbf{Question Distribution}. \figref{fig:comp_ques} shows the distribution of questions based on the first four words of the questions for the train (left) and test (right) splits of the C-VQA dataset. We can see that splitting the dataset compositionally (as in C-VQA) does not lead to significant differences in the distribution of questions across splits, keeping the distributions qualitatively similar to VQA v1.0 splits \cite{VQA}. Quantitatively, 46.06\% of the question strings in the VQA v1.0 val split are also present in the VQA v1.0 train split, whereas this percentage is 37.76 for the C-VQA splits.

% \begin{figure*}[h]
% \centering
% \includegraphics[width=1\linewidth]{figures/orig_ques.png}
% \caption{Distribution of questions by their first four words for a random sample of 60K questions for \textbf{VQA v1.0 train split} (left) and \textbf{VQA v1.0 val split} (right). The ordering of the words starts towards the center and radiates outwards. The arc length is proportional to the number of questions containing the word. White areas are words with contributions too small to show.}
% %\vspace{-5pt}
% \label{fig:orig_ques}
% %\setlength{\belowcaptionskip}{-10pt}
% \end{figure*}

\begin{figure*}[h]
\centering
\includegraphics[width=1\linewidth]{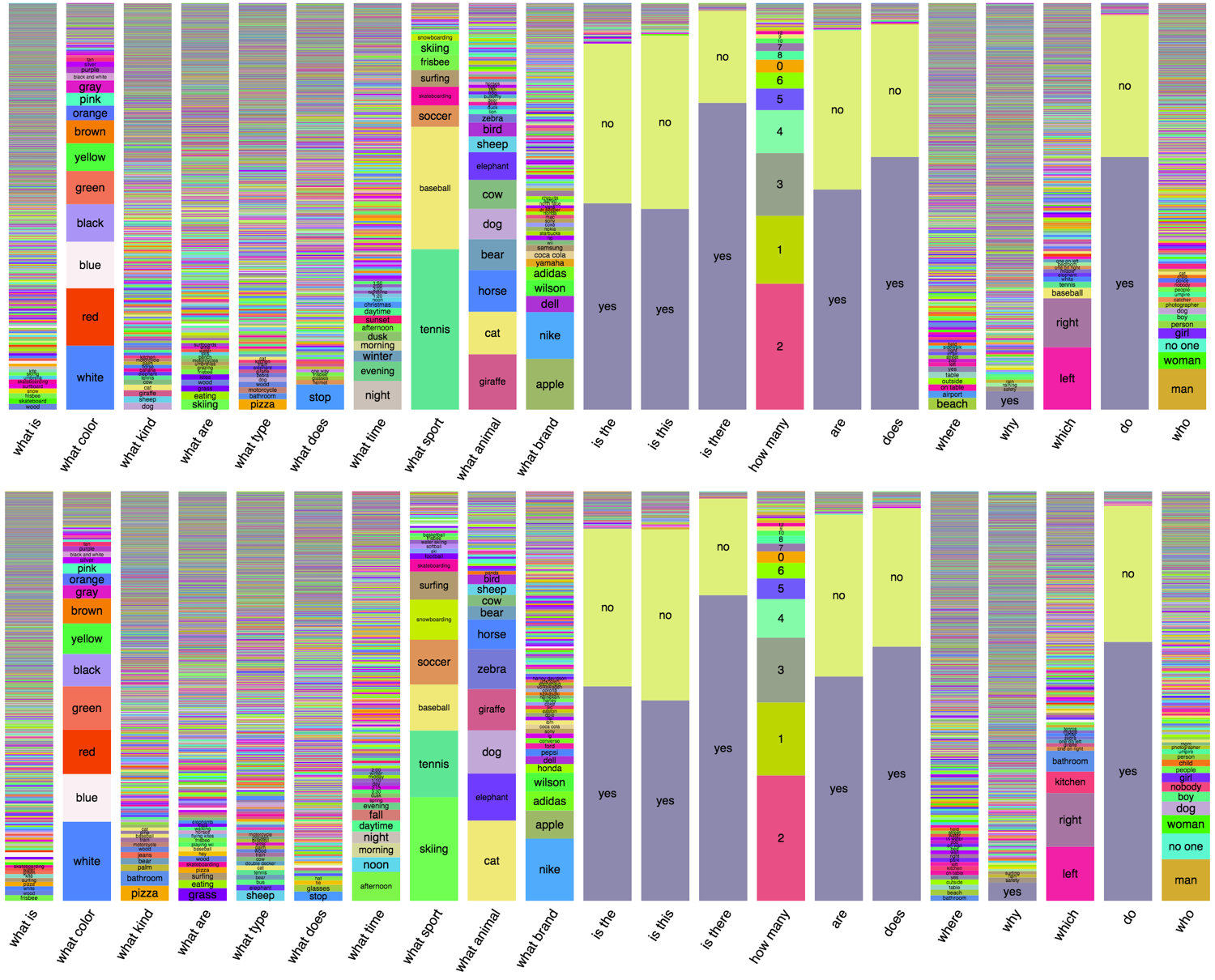}
\caption{Distribution of answers per question type for a random sample of 60K questions for \textbf{C-VQA train split} (top) and \textbf{C-VQA test split} (bottom).}
%\vspace{-5pt}
\label{fig:comp_ans}
\end{figure*}

\textbf{Answer Distribution.} \figref{fig:comp_ans} shows the distribution of answers for several question types such as ``what color'', ``what sport'', ``how many'', etc. for the train (top) and test (bottom) splits of the C-VQA dataset. We can see that the distributions of answers for a given question type is significantly different. However, for VQA v1.0 dataset, the distribution for a given question type is similar across train and val splits \cite{VQA}.  For instance, ``tennis'' is the most frequent answer for the question type ``what sport'' in C-VQA train split whereas ``skiing'' is the most frequent answer for the same question type in C-VQA test split. However, for the VQA v1.0 splits, ``tennis'' is the most frequent answer for both the train and val splits. Similar differences can be seen for other question types as well -- ``what animal'', ``what brand'', ``what kind'', ``what type'', ``what are''. Quantitatively, 32.49\% of the QA pairs in the VQA v1.0 val split are also present in the VQA v1.0 train split, whereas this percentage is 0 for the C-VQA splits (by construction).

% \begin{figure*}[h]
% \centering
% \includegraphics[width=1\linewidth]{figures/orig_ans.png}
% \caption{Distribution of answers per question type for a random sample of 60K questions for \textbf{VQA v1.0 train split} (top) and \textbf{VQA v1.0 val split} (bottom).}
% %\vspace{-5pt}
% \label{fig:orig_ans}
% %\setlength{\belowcaptionskip}{-10pt}
% \end{figure*}

\section{Baselines}
\label{sec:baselines}

We report the performances of the following VQA models when trained on C-VQA train split and evaluated on C-VQA test split and compare this with the setting when these models are trained on VQA v1.0 train split and evaluated on VQA v1.0 val split (\tableref{table:baselines}).

\textbf{Deeper LSTM Question + normalized Image (deeper LSTM Q + norm I)} \cite{Lu2015}: This model was proposed in \cite{VQA}. It is a two channel model -- one channel processes the image and the other channel processes the question. For each image, the image channel extracts the activations (4096-dim) of the last hidden layer of the VGGNet~\cite{VGG} and normalizes them. For each question, the question channel extracts the hidden state and cell state activations of the last hidden layers of 2-layered LSTM, resulting in a 2048-dim encoding of the question. The image features (4096-dim) obtained from the image channel and the question features (2048-dim) obtained from the question channel are linearly transformed to 1024 dimensions each and fused together via element-wise multiplication. This fused vector is then passed through one more fully-connected layer in a Multi-Layered Perceptron (MLP), which finally outputs a 1000-way softmax score over the 1000 most frequent answers from the training set. The entire model, except the CNN (which is not fine-tuned) is learned end-to-end with a cross-entropy loss.

\textbf{Neural Module Networks (NMN)} \cite{DBLP:journals/corr/AndreasRDK15}: This model is designed to be compositional in nature. The model consists of composable modules where each module has a specific role (such as detecting a dog in the image, counting the number of dogs in the image, etc.). Given an image and the natural language question about the image, NMN first decomposes the question into its linguistic substructures using a parser. These structures determine which modules need to be composed together in what layout to create the network for answering the question. The resulting compound networks are jointly trained. At test time, the image and the question are forward propagated through the dynamically composed network which outputs a distribution over answers. In addition to the network composed using different modules, NMN also uses an LSTM to encode the question which is then added elementwise to the representation produced by the last module of the NMN. This combined representation is passed through a fully-connected layer to output a softmax distribution over answers. The LSTM encodes priors in the training data and models syntactic regularities such as singular vs. plural (``what is flying?'' should be answered with ``kite'' whereas ``what are flying?'' should be answered with ``kites'').

\textbf{Stacked Attention Networks (SAN)} \cite{DBLP:journals/corr/YangHGDS15}: This is one of the widely used models for VQA. This model is different from other VQA models in that it uses multiple hops of attention over the image. Given an image and the natural language question, SAN uses the question to obtain an attention map over the image. The attended image is combined with the encoded question vector which becomes the new query vector. This new query vector is used again to obtain a second round of attention over the image. The query vector obtained from the second round of attention is passed through a fully-connected layer to obtain a distribution over answers.\footnote{We use a torch implementation of SAN, available at \url{https://github.com/abhshkdz/neural-vqa-attention}, for our experiments.}

\textbf{Hierarchical Question-Image Co-attention Networks (HieCoAtt)} \cite{hieco}: This is one of the top performing models for VQA. In addition to modeling attention over image, this model also models attention over question. Both image and question attention are computed in a hierarchical fashion. The attended image and question features obtained from different levels of the hierarchy are combined and passed through a fully-connected layer to obtain a softmax distribution over the space of answers.

\textbf{Multimodal Compact Bilinear Pooling (MCB)} \cite{fukui}: This model won the real image track of the VQA Challenge 2016. MCB uses multimodal compact bilinear pooling to predict attention over image features and also to combine the attended image features with the question features. These combined features are passed through a fully-connected layer to obtain a softmax distribution over the space of answers.

\begin{table}[h]
\setlength{\tabcolsep}{3.2pt}
{\small
\begin{center}
\begin{tabular}{@{}llcccc@{}}
\toprule
Model & Dataset & Yes/No & Number & Other & Overall\\
%\hline
\midrule
    & VQA v1.0 val & 79.81 & 33.26 & 40.35 & 54.23\\
deeper LSTM Q + norm I \cite{Lu2015}  &  &  &  & \\
    & C-VQA test & 70.60 & 29.76 & 31.83 & 46.69\\
%\hline
% \midrule
%     & VQA v1.0 val & 64.91 & 11.89 & 09.96 & 30.77 \\
% Neural Module Networks \cite{DBLP:journals/corr/AndreasRDK15}  &  &  &  & \\
%     & C-VQA test & 62.40 & 13.46 & 07.63 & 29.70 \\
% %\hline
\midrule
    & VQA v1.0 val & 80.39 & 33.45 & 41.07 & 54.83 \\
NMN \cite{DBLP:journals/corr/AndreasRDK15}  &  &  &  & \\
    & C-VQA test & 72.96 & 31.02 & 34.49 & 49.05 \\
%\hline
\midrule
    & VQA v1.0 val & 78.54 & 33.46 & 44.51 & 55.86\\
SAN \cite{DBLP:journals/corr/YangHGDS15} &  &  &  & \\
    & C-VQA test & 66.96 & 24.30 & 33.19 & 45.25 \\
%\hline
\midrule
    & VQA v1.0 val & 79.81 & 34.93 & 45.64 & 57.09 \\
HieCoAtt \cite{hieco}  &  &  &  & \\
    & C-VQA test & 71.11 & 30.48 & 38.31 & 50.12 \\
%\hline
\midrule
    & VQA v1.0 val & 81.62 & 34.56 & 52.16 & 60.97 \\
MCB \cite{fukui}  &  &  &  & \\
    & C-VQA test & 71.33 & 24.90 & 47.84 & 54.15 \\
%\hline
\bottomrule
\end{tabular}
\end{center}
}
\caption {Accuracies of existing VQA models on the VQA v1.0 val split when trained on VQA v1.0 train split and those on C-VQA test split when trained on C-VQA train split.}
\label{table:baselines}
\end{table}

From \tableref{table:baselines}, we can see that the performance of all the existing VQA models drops significantly in the C-VQA setting compared to the VQA v1.0 setting. Note that even though the Neural Module Networks architecture is compositional by design, their performance suffers on C-VQA. We posit this may be because they use an additional LSTM encoding of the question to encode priors in the dataset. In C-VQA, the priors learned from the train set are unlikely to generalize to the test set. Also note that other models suffer a larger drop in performance compared to Neural Module Networks. 

Another interesting observation from \tableref{table:baselines} is that the ranking of the models based on overall performance changes from VQA v1.0 to C-VQA. For VQA v1.0, SAN outperforms deeper LSTM Q + norm I and NMN, whereas for C-VQA, these two models outperform SAN. Also note the change in ranking of the models for different types of answers (``yes/no'', ``number'', ``other''). For instance, for ``number'' questions, MCB outperforms all the models except HieCoAtt for VQA v1.0. However, for C-VQA, all the models except SAN outperform MCB. 

Examining the accuracies of these models for different question types shows that the performance drop from VQA v1.0 to C-VQA is larger for some question types than the others. For Neural Module Networks (NMN), Stacked Attention Networks (SAN) and Hierarchical Question-Image Co-attention Networks (HieCoAtt), questions starting with ``what room is'' (such as ``What room is this?'') have the largest drop -- 33.28\% drop for NMN, 40.73\% drop for SAN and 32.56\% drop for HieCoAtt. For such questions in the C-VQA test split, one of the correct answers is ``living room'' which is not one of the correct answers to such questions in the C-VQA train split (the correct answers in the C-VQA train split are ``kitchen'', ``bedroom'', etc.). So, models tend to answer the C-VQA test questions with what they have seen during training (such as ``kitchen''). Note that ``living room'' is seen during training for questions such as ``Which room is this?''. For deeper LSTM + norm I model and Multimodal Compact Bilinear Pooling (MCB) model, the largest drop is for ``is it'' questions (such as ``Is it daytime?'') -- 29.52\% drop for deeper LSTM Q + norm I and 30.77\% drop for MCB model. For such questions in the C-VQA test split, the correct answer is ``yes'' whereas the correct answer for such questions in C-VQA train split is ``no''. Again, models tend to answer the C-VQA test questions with ``no''. Other question types resulting in significant drop in performance (more than 10\%) for all the models are -- ``what is the color of the'', ``how many people are in'', ``are there'', ``is this a''. 

% ``what color is'', ``what is the man''
% \input{model}
% \input{experiments}
\section{Conclusion}
\label{sec:conclusion}

In conclusion, we introduce a novel setting for Visual Question Answering -- Compositional Visual Question Answering. Under this setting, the question-answer pairs in the test set are compositionally novel compared to the question-answer pairs in the training set. We create a compositional split of the VQA (v1.0) dataset \cite{VQA}, called C-VQA, which facilitates training compositional VQA models. We show the similarities and differences between the VQA v1.0 and C-VQA splits. Finally, we report performances of several existing VQA models on the C-VQA splits and show that the performance of all the models drops significantly compared to the original VQA v1.0 setting. This suggests that today's VQA models do not handle compositionality well and that C-VQA splits can be used as a benchmark for building and evaluating compositional VQA models. 

{\small
\bibliographystyle{unsrt}
\bibliography{references}
}

\end{document}